\documentclass{article}

\usepackage[preprint]{icml2026}

\usepackage[utf8]{inputenc}
\usepackage[T1]{fontenc}

\usepackage[font=small]{caption}
\usepackage{xcolor}
\definecolor{cornflowerblue}{rgb}{0.39, 0.58, 0.93}

\usepackage{hyperref}
\hypersetup{
  colorlinks=true,
  linkcolor=cornflowerblue,
  filecolor=magenta,
  urlcolor=teal,
  citecolor=cornflowerblue,
  pdftitle={},
  pdfpagemode=FullScreen
}

\usepackage{url}
\usepackage{microtype}
\usepackage{graphicx}
\usepackage{wrapfig}
\usepackage{booktabs}
\usepackage{multirow}
\usepackage{placeins}

\usepackage{amsmath,amssymb}
\usepackage{amsfonts}
\usepackage{amsthm}
\usepackage{nicefrac}
\usepackage{bbm}

\usepackage{booktabs}
\usepackage{amsmath} 
\usepackage{graphicx} 

\usepackage{enumitem}
\usepackage{xspace}
\usepackage[capitalize,noabbrev]{cleveref}

\usepackage{tikz}
\usetikzlibrary{positioning, shapes.geometric, arrows.meta, calc, fit}
\usepackage[table]{xcolor}
\usepackage{algorithm}
\usepackage{algorithmic}

\usepackage{footmisc}
\usepackage{lineno}

\theoremstyle{plain}

\theoremstyle{definition}

\theoremstyle{remark}

\usepackage{listings}
\definecolor{codebg}{gray}{0.96}
\definecolor{codeframe}{gray}{0.80}
\definecolor{pykeyword}{RGB}{0,0,180}
\definecolor{pycomment}{RGB}{0,120,0}
\definecolor{pystring}{RGB}{180,0,0}
\definecolor{pynumber}{RGB}{120,0,120}

\lstdefinestyle{py}{
  language=Python,
  basicstyle=\ttfamily\footnotesize,
  keywordstyle=\color{pykeyword}\bfseries,
  commentstyle=\color{pycomment}\itshape,
  stringstyle=\color{pystring},
  numberstyle=\tiny\color{pynumber},
  showstringspaces=false,
  breaklines=true,
  breakatwhitespace=true,
  keepspaces=false,
  columns=fullflexible,
  frame=single,
  rulecolor=\color{codeframe},
  backgroundcolor=\color{codebg},
  tabsize=2,
  xleftmargin=2pt,
  framexleftmargin=2pt,
  aboveskip=0.4em,
  belowskip=0.4em
}
\icmltitlerunning{Speculating Experts Accelerates Inference for Mixture-of-Experts}

\begin{document}

\twocolumn[
  \icmltitle{Speculating Experts Accelerates Inference for Mixture-of-Experts}

  \begin{icmlauthorlist}
    \icmlauthor{Vivan Madan\(^{*}\)}{umd}
    \icmlauthor{Prajwal Singhania\(^{*}\)}{umd}
    \icmlauthor{Abhinav Bhatele}{umd}
    \icmlauthor{Tom Goldstein}{umd}
     \icmlauthor{Ashwinee Panda}{umd,togetherai}
   \end{icmlauthorlist}
   \icmlaffiliation{umd}{University of Maryland}
   \icmlaffiliation{togetherai}{TogetherAI}
   \icmlcorrespondingauthor{Ashwinee Panda}{ashwinee@umd.edu}
]

\printAffiliationsAndNotice{\icmlEqualContribution}

\begin{figure}[h]
    \centering
    \includegraphics[scale=0.5]{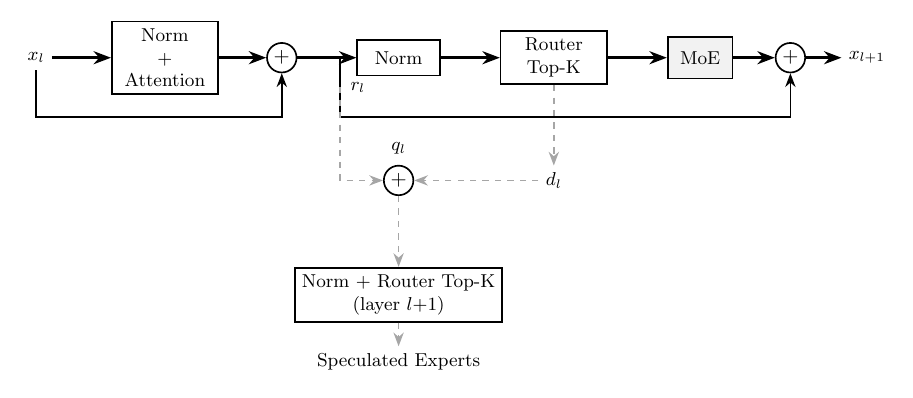}
    \caption{Expert prefetching in a pre-norm MoE block. The normalized residual stream $s_l$ and default vector $d_l$ at layer $l$ form the quasi-hidden state $q_l$ which is used to predict the next-layer's experts, enabling CPU-GPU memory transfer to overlap with computation.}
    \label{fig:moe_prefetch_graphic}
\end{figure}

\begin{abstract}
Mixture-of-Experts (MoE) models have gained popularity as a means of scaling the capacity of large language models (LLMs) while maintaining sparse activations and reduced per-token compute. However, in memory-constrained inference settings, expert weights must be offloaded to CPU, creating a performance bottleneck from CPU-GPU transfers during decoding. We propose an expert prefetching scheme that leverages currently computed internal model representations to speculate future experts, enabling memory transfers to overlap with computation. Across multiple MoE architectures, we demonstrate that future experts can be reliably predicted by these internal representations. We also demonstrate that executing speculated experts generally maintains downstream task accuracy, thus preserving more effective compute-memory overlap by eliminating the need to re-fetch true router-selected experts. Integrated into an optimized inference engine, our approach achieves up to 14\% reduction in time per output token (TPOT) over on-demand loading of experts from CPU memory. For MoEs where speculative execution alone yields suboptimal accuracy, we further examine lightweight estimators that improve expert prediction hit rates, thereby reducing performance degradation. Our code is released in open-source at \href{https://github.com/axonn-ai/yalis/tree/offload_prefetch}{https://github.com/axonn-ai/yalis/tree/offload\_prefetch}.
\end{abstract}
\begin{figure}[t]
    \centering
    \includegraphics[width=\columnwidth]{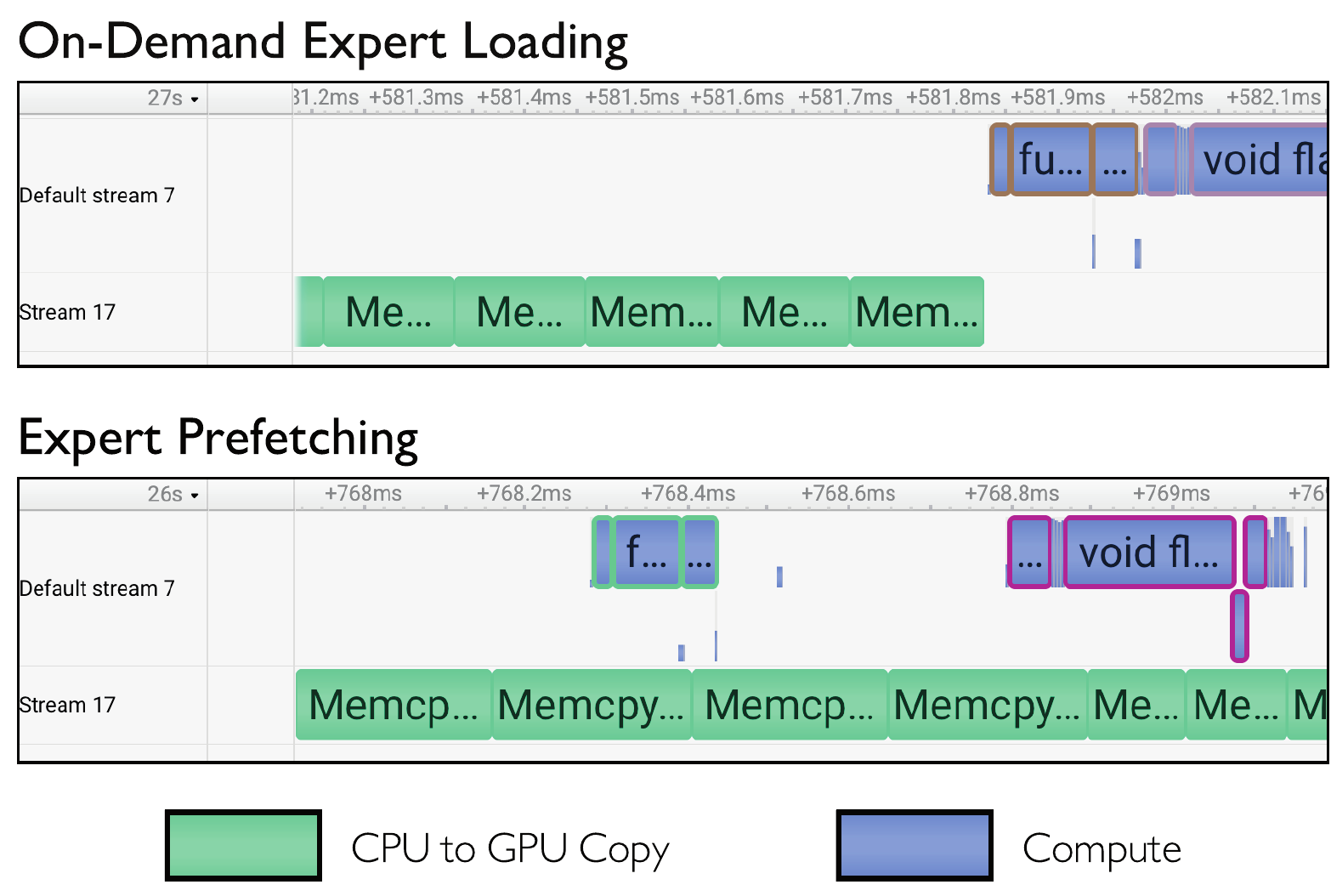}
    \caption{Nsight Systems trace for inference with Qwen-30B-A3B. Compared to on-demand loading of active experts (top), our expert prefetching approach (bottom) effectively overlaps CPU-GPU memory transfers with GPU computation, reducing transfer overhead on the critical path.}
    \label{fig:trace}
\end{figure}

\section{Introduction}

Mixture-of-Experts (MoE) architectures have emerged as an effective approach for scaling large language models (LLMs) by increasing parameter count without a proportional increase in per-token computation. MoEs replace dense activations in feed-forward layers with sparsely activated experts selected by a learned gating function \cite{large_nns}. As a result, MoEs have become the predominant architecture for many state-of-the-art LLMs such as Qwen3-MoE, GPT-OSS, and GLM 4.7 \cite{yany25_qwen3_tech_report, openai25_oss_model_card, glm45_25_tech_report}. This sparse activation makes MoEs attractive for deployment in resource-constrained environments where large dense models may be infeasible due to limited GPU memory. In such settings, the majority of expert parameters used in MoEs can be offloaded to CPU RAM with only a small subset of these experts being transferred and executed on-device. However, inference often becomes I/O-bound, with CPU-GPU transfer latency dominating time per output token (TPOT), and significantly reducing throughput \cite{chen25_spmoe, yu25_pre_scope, zhang25_duoserve_moe_expert_pf}. For example, for Qwen3-30B-A3B on an A6000 GPU, memory transfers from CPU-GPU account for ~84-88\% of the TPOT while compute accounts for much less. So, significant gains in TPOT can be achieved in inference speeds by reducing these expensive memory transfers.

Efficient inference of MoEs is important for enabling broader access and control over many state-of-the-art open-source LLMs. In this work, we introduce an inference-time expert prefetching scheme that leverages internal model representations to predict near-future expert selection, operates directly on existing pretrained MoE models, and is ready-to-use in an open-source inference engine YALIS \cite{yalis}. The prefetching scheme transfers future layer experts in parallel with computation, leading to improved inference latencies. 



Our contributions are as follows:
\begin{itemize}
    \item \textbf{Parameter-free prefetching.} We identify internal model representations that contain signals capable of predicting future routing decisions across modern MoE architectures with large expert pools.

    \item \textbf{Speculative execution preserving accuracy.} We demonstrate that executing prefetched experts, rather than treating mispredictions as cache misses, often maintains downstream task accuracy.

    \item \textbf{An optimized inference implementation.}
    We integrate our prefetching scheme into an optimized open-source inference engine and demonstrate a 5-14\% reduction in time per output token (TPOT) over on-demand CPU expert loading across hardware/model configurations in resource-constrained settings. Our code is released in open-source at \href{https://github.com/axonn-ai/yalis/tree/offload_prefetch}{https://github.com/axonn-ai/yalis/tree/offload\_prefetch}.

    \item \textbf{Lightweight neural estimators.} For architectures where the prefetching scheme falls short due to large representation drift over certain layers, we introduce a lightweight neural estimator that substantially improves expert prediction hit rates over those layers while requiring only a small number of training tokens.
    
\end{itemize}

\section{Related Work}

To address the inference bottleneck caused by CPU-GPU transfer latency, many prior works implement \textbf{expert caching} where popular experts are kept on-device to remove redundant transfers. Other approaches employ \textbf{expert prefetching}, in which experts predicted to be used in future layers are transferred to the GPU and overlapped with the current layer's computation, masking CPU-GPU transfer latency. Although effective in reported settings, Zhang et al. evaluate on MoE architectures with relatively small routing combinatorial spaces, leaving it unclear whether such predictor-based schemes extend to modern MoEs with larger expert pools \cite{zhang25_duoserve_moe_expert_pf}. Chen et al. leverage internal model representations from a draft model to speculate experts that may be used in the larger model \cite{chen25_spmoe}. Yu et al. utilize both a layer-wise predictor and a cross-layer scheduler to prefetch experts multiple layers in advance \cite{yu25_pre_scope}.  Hwang et al. modify MoEs with a pre-gate function that decouples expert selection and execution, enabling one-layer-in-advance prefetching, but requires additional full-model fine-tuning \cite{hwang24_pregated_moe}. Yu et al. maintain expert maps that contain input semantic embeddings and expert activation trajectories to predict future expert use \cite{yu25_taming_latency_mem_tradeoff_exp_offloading}. Eliseev et al. use LRU-based expert caching and apply the next-layer's gating function to the current layer's hidden states to speculate future experts \cite{eliseev_fast_inference}. Zhang et al. modify part of the standard LLM architecture by rerouting the residual stream, allowing current AllReduce communication operations used in tensor parallelism settings to overlap with subsequent MLP or attention computation \cite{zhang25_ladder_res}. 

Several of these prior approaches leverage internal representations for future expert prediction; however, the predictions are treated as cache hints, and model execution still follows the true expert utilization. As a result, prediction error forces re-fetching of missed experts, limiting the extent to which computation and memory transfer can be overlapped. While we do utilize similar internal representations, we further enhance them by adding an additional expert-conditioned signal to improve expert hit rate and study \emph{speculative execution} where predicted experts selected by the next layer's router are utilized, thus allowing for better preservation of compute-memory overlap in modern MoE architectures. We analyze the impact of speculative execution on downstream accuracy and show that a lightweight estimator can help mitigate performance degradation for certain model architectures and their few high-drift layers where router-based speculation is insufficient.


\section{Signals for Expert Prefetching}

\subsection{Default Vector}
\citet{panda2025densebackpropagationimprovestraining} introduce the \textit{default vector}, denoted $d_{l, e}$, that represents the average activation associated with expert $e$ at layer $l$. These vectors are computed offline by aggregating the observed activations for each selected expert during inference.

Given a token at layer $l$, the router selects a subset of experts $\mathcal{E}_l$ (commonly $\text{Top-K}$) and assigns the corresponding gating weights $\mathcal{G}_l = \{g_{l, e}\}_{e \in \mathcal{E}_l}$. We define the layer-level default representation $d_l$ as the weighted combination of the default vectors of the selected experts and their assigned weight:
\[
d_l = \sum_{e \in \mathcal{E}_l} g_{l, e}d_{l, e}
\]

Effectively, $d_l$ captures the expert-conditioned typical contribution to the residual stream by an MoE block.

\subsection{Quasi-Hidden State}
In the pre-norm MoE architectures considered in this work, expert routing is computed from a normalized residual stream following attention. Abiding by this structure, we define the quasi-hidden state as an approximate to the router input.

Specifically, let $r_l$ denote the post-attention residual at layer $l$, and let $d_l$ denote the layer-level default vector. The quasi-hidden state $q_l$ is defined as
\[
\text{q}_l = \mathrm{LN}_{l+1}(d_l + r_l) 
\]
where $\mathrm{LN}_{l+1}(\cdot)$ denotes the normalization applied to the residual stream prior to expert routing at layer $l+1$.

\subsection{Prefetching Over Representations}

\begin{figure}[h]
    \centering
    \includegraphics[width=0.9\linewidth]{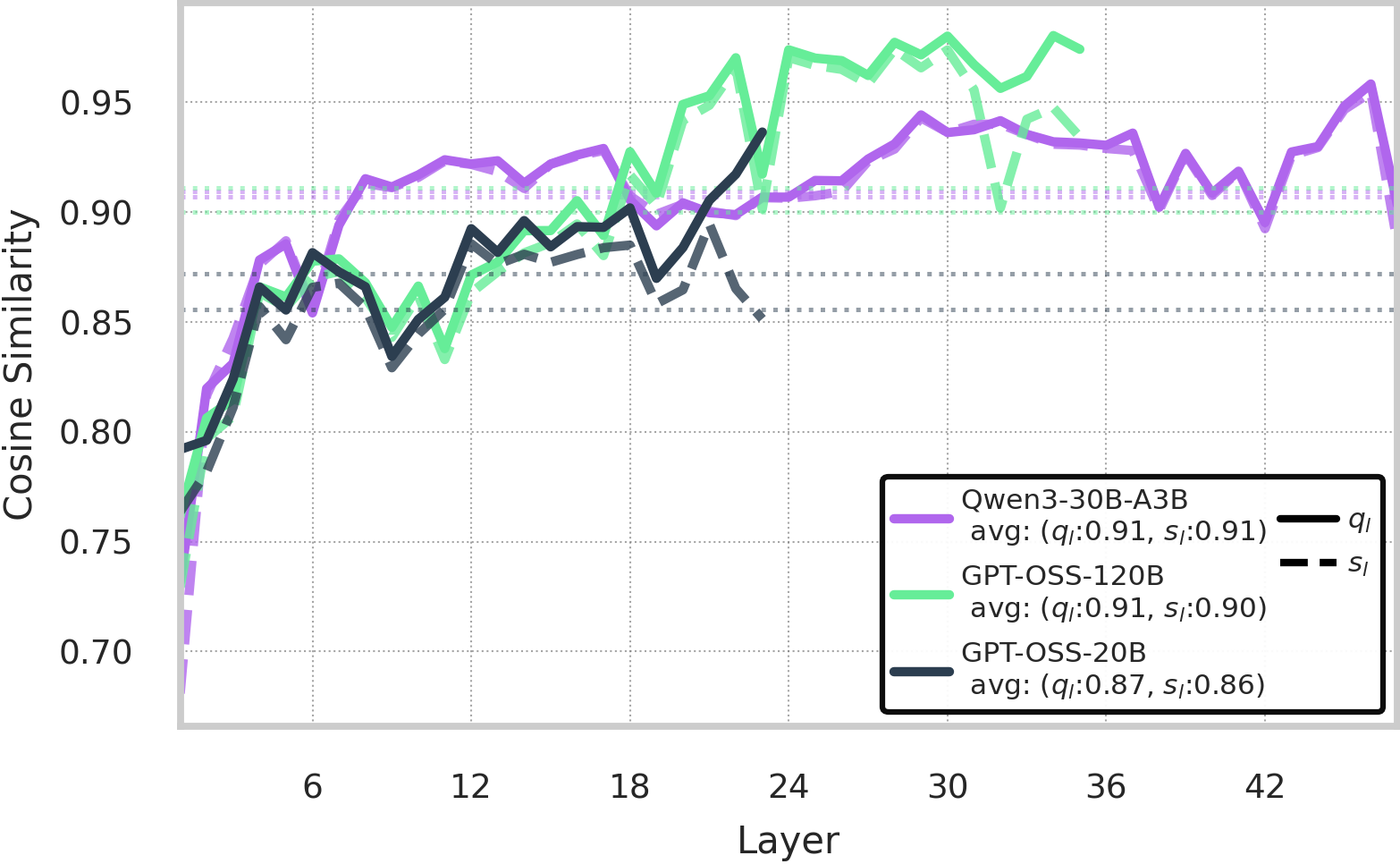}    
    \caption{Comparison of cosine similarity between the quasi-hidden state $q_l$ constructed at layer $l$ and the true router input $s_{l+1}$.}
    \label{fig:sim}
\end{figure}

\begin{figure}[h]
    \centering
    \includegraphics[width=0.9\linewidth]{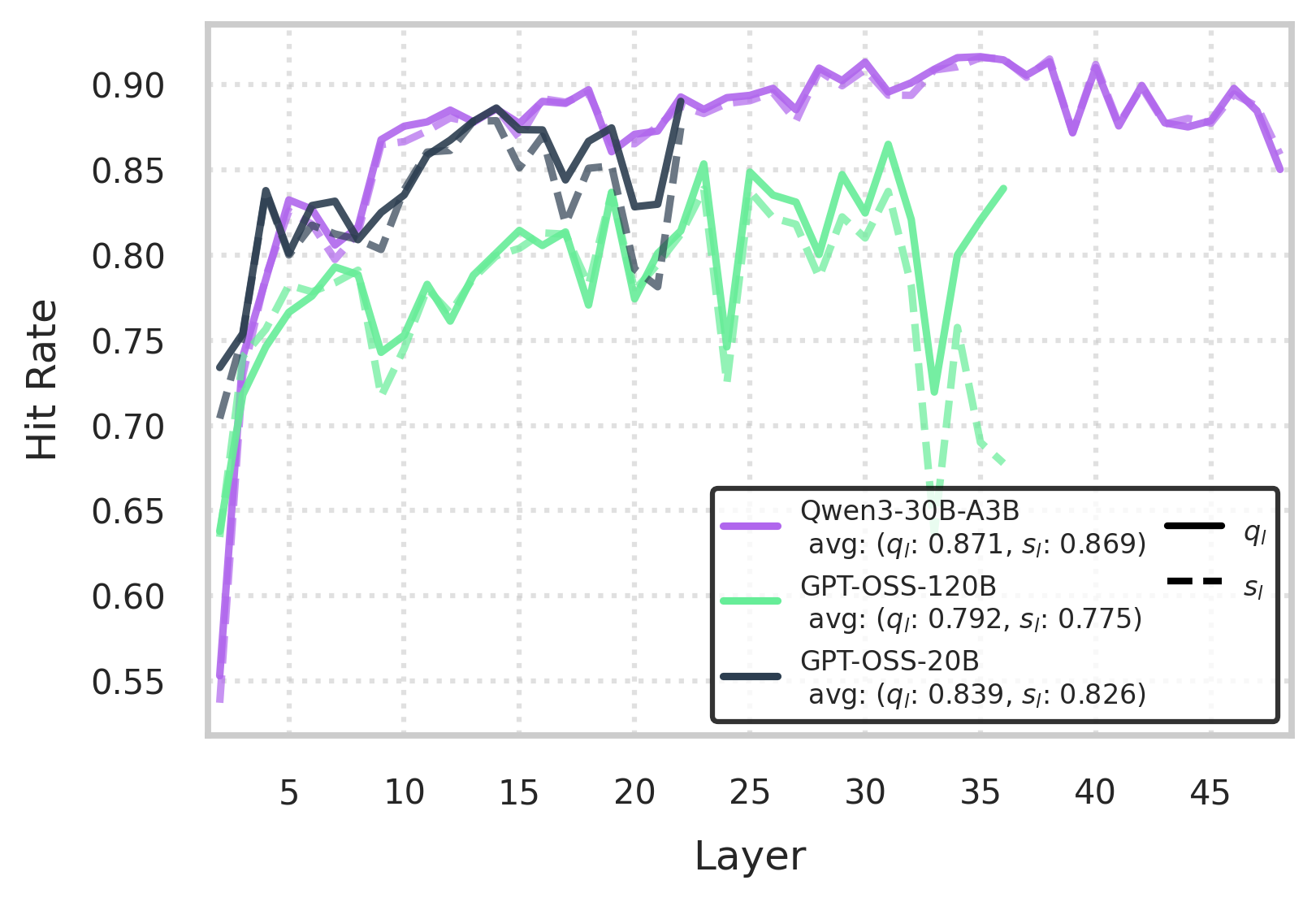}
    \caption{Per-layer expert prefetch hit rates (recall@k) obtained using the quasi-hidden state $q_l$ versus the baseline $s_l$ across models.}
    \label{fig:hit_rates}
\end{figure}

We evaluate two internal representations as candidate inputs for expert prefetching at layer $l+1$: the normalized residual stream $s_{l}$ and the quasi-hidden state $q_l$.

The representation $s_l$ corresponds to the normalized residual stream produced at layer $l$ and directly fed into the router. As such, $s_l$ serves as a natural baseline, representing the exact signal used for routing in the current layer.

We measure the cosine similarity in Figure~\ref{fig:sim} between each representation constructed at layer $l$ and the ground-truth input $s_{l+1}$. Across the GPT-OSS models, the quasi-hidden states demonstrate higher average cosine similarity than $s_l$, indicating that the inclusion of the default vectors provides a useful expert-conditioned bias for approximating drift between $s_l$ and $s_{l+1}$.

In contrast, for Qwen3-30B-A3B, quasi-hidden states yield only negligible improvements over $s_l$. In this model, the drift between $s_l$ and $s_{l+1}$ is substantially smaller beyond the early layers, reducing the benefit of incorporating the default vector. 

In Figure~\ref{fig:hit_rates}, recall@k (where k is the number of active experts) between the predicted and ground-truth experts further supports the hypothesis that improved alignment with the true routing signal reduces expert selection error. In particular, representations with higher cosine similarity to the true router input at layer $l+1$ achieve higher recall@k. 

For Qwen3-30B-A3B, most of the inter-layer drift occurs within the first two layers, corresponding to lower recall@k in this early regime. Beyond these layers, the negligible drift allows for a much higher recall@k of approximately $90\%$ on average. For the GPT-OSS models, the quasi-hidden states improve relative error rate substantially over the baseline where drift is much more prominent, notably in the beginning and end layers. All results in this section assume the router-based prefetching depicted in Figure~\ref{fig:moe_prefetch_graphic}.




\section{Effectiveness of Prefetched Experts}
\label{sec:effec_pfing}

In prior prefetching approaches, mispredicted experts are ignored during execution, and router-selected experts that are absent from GPU memory are loaded on demand as cache misses \cite{yu25_taming_latency_mem_tradeoff_exp_offloading, zhang25_duoserve_moe_expert_pf, xue2025_moe_infinity}. Although this scheme maintains correctness, it reduces the degree to which computation can overlap with expert loading, limiting the practical benefits of prefetching. 

In this section, we evaluate the effect of \emph{utilizing} all prefetched experts during inference. Specifically, we execute predicted experts and their associated routing weights rather than treating them as cache misses. We demonstrate that this approach preserves downstream task accuracy across a range of reasoning-heavy benchmarks. 

\subsection{Prefetched Experts on Task Accuracy}

\begin{table*}[t]
\centering
\caption{Downstream benchmark accuracy $\pm$ standard error. Grey shading denotes our proposed expert-prefetching schemes. We bold the \textbf{best} and underline the \underline{second-best} performing variant within each model family.}
\label{tab:model_performance}

\renewcommand{\arraystretch}{1.2}
\setlength{\tabcolsep}{8pt}
\definecolor{lightgrey}{gray}{0.92}

\resizebox{\textwidth}{!}{%
\begin{tabular}{@{}l|cccccc|c@{}}
\toprule
 & \textbf{HumanEval} & \textbf{MBPP+} & \textbf{GSM8k} & \textbf{AIME24} & \textbf{AIME25} & \textbf{StrategyQA} & \textbf{Average} \\
\midrule
Qwen3-30B-A3B & $\mathbf{0.939}_{\pm 0.019}$ & $\mathbf{0.762}_{\pm 0.022}$ & $\mathbf{0.950}_{\pm 0.006}$ & $\mathbf{0.800}_{\pm 0.073}$ & $\mathbf{0.733}_{\pm 0.081}$ & $\mathbf{0.719}_{\pm 0.017}$ & $\mathbf{0.817}$ \\
\rowcolor{lightgrey} Qwen3-30B-A3B + Router-PF & $0.860_{\pm 0.027}$ & $0.659_{\pm 0.024}$ & $0.576_{\pm 0.014}$ & $0.467_{\pm 0.091}$ & $0.600_{\pm 0.089}$ & $0.683_{\pm 0.018}$ & $0.641$ \\
\rowcolor{lightgrey} Qwen3-30B-A3B + Est-PF & $\underline{0.915}_{\pm 0.022}$ & $0.741_{\pm 0.023}$ & $0.918_{\pm 0.008}$ & $0.667_{\pm 0.086}$ & $0.567_{\pm 0.090}$ & $\underline{0.691}_{\pm 0.018}$ & $0.750$ \\
\rowcolor{lightgrey} Qwen3-30B-A3B + Hybrid-PF & $0.909_{\pm 0.023}$ & $\underline{0.762}_{\pm 0.022}$ & $\underline{0.946}_{\pm 0.006}$ & $\underline{0.700}_{\pm 0.084}$ & $\underline{0.600}_{\pm 0.089}$ & $0.681_{\pm 0.018}$ & $\underline{0.766}$ \\
\midrule
GPT-OSS-20B & $\mathbf{0.970}_{\pm 0.013}$ & $\underline{0.794}_{\pm 0.021}$ & $\mathbf{0.942}_{\pm 0.006}$ & $\mathbf{0.667}_{\pm 0.086}$ & $\mathbf{0.667_{\pm 0.086}}$ & $\mathbf{0.753}_{\pm 0.016}$ & $\mathbf{0.799}$ \\
\rowcolor{lightgrey} GPT-OSS-20B + Router-PF & $\underline{0.933}_{\pm 0.020}$ & $\mathbf{0.804}_{\pm 0.020}$ & $0.929_{\pm 0.007}$ & $\mathbf{0.667}_{\pm 0.086}$ & $\underline{0.600}_{\pm 0.089}$ & $\underline{0.738}_{\pm 0.017}$ & $\underline{0.779}$ \\
\rowcolor{lightgrey} GPT-OSS-20B + Est-PF & $0.884_{\pm 0.025}$ & $0.751_{\pm 0.022}$ & $\underline{0.933}_{\pm 0.007}$ & $0.633_{\pm 0.088}$ & $0.533_{\pm 0.091}$ & $0.726_{\pm 0.017}$ & $0.743$ \\
\rowcolor{lightgrey} GPT-OSS-20B + Hybrid-PF & $0.896_{\pm 0.024}$ & $0.788_{\pm 0.021}$ & $0.936_{\pm 0.007}$ & $\mathbf{0.667}_{\pm 0.086}$ & $0.467_{\pm 0.091}$ & $0.725_{\pm 0.017}$ & $0.747$ \\
\midrule
GPT-OSS-120B & $\mathbf{0.970}_{\pm 0.013}$ & $\mathbf{0.815}_{\pm 0.020}$ & $0.955_{\pm 0.006}$ & $\underline{0.800}_{\pm 0.073}$ & $\underline{0.767}_{\pm 0.077}$ & $\mathbf{0.789}_{\pm 0.016}$ & $\underline{0.849}$ \\
\rowcolor{lightgrey} GPT-OSS-120B + Router-PF & $\underline{0.963}_{\pm 0.015}$ & $\underline{0.812}_{\pm 0.020}$ & $\mathbf{0.958}_{\pm 0.006}$ & $\mathbf{0.833}_{\pm 0.068}$ & $\mathbf{0.800}_{\pm 0.073}$ & $\underline{0.776}_{\pm 0.016}$ & $\mathbf{0.857}$ \\
\rowcolor{lightgrey} GPT-OSS-120B + Est-PF & $0.951_{\pm 0.017}$ & $0.804_{\pm 0.020}$ & $0.949_{\pm 0.006}$ & $0.700_{\pm 0.084}$ & $0.700_{\pm 0.084}$ & $0.760_{\pm 0.016}$ & $0.811$ \\
\rowcolor{lightgrey} GPT-OSS-120B + Hybrid-PF & $0.927_{\pm 0.020}$ & $0.802_{\pm 0.021}$ & $\underline{0.945}_{\pm 0.006}$ & $\underline{0.800}_{\pm 0.073}$ & $\mathbf{0.800}_{\pm 0.073}$ & $0.769_{\pm 0.016}$ & $0.841$ \\
\bottomrule
\end{tabular}%
}
\end{table*}

Table~\ref{tab:model_performance} reports these downstream task accuracies for router-based prefetching (\textbf{Router-PF}) on a variety of benchmarks that cover coding, math, and common sense reasoning. For both GPT-OSS models, executing prefetched experts for layer $l+1$ using the quasi-hidden states constructed at layer $l$ preserves performance over most tasks relative to baseline performance. This suggests that for those architectures, internal representations encode signals that are sufficiently predictive of future expert selection, preserving downstream task performance. 

Qwen3-30B-A3B exhibits more sensitivity to router-based speculative execution.  In particular, common sense reasoning (StrategyQA) and simple coding tasks (HumanEval) incur less drastic performance degradation than math-heavy tasks such as AIME24 and GSM8k. This reduced task accuracy can be attributed to the high representational drift observed in the early layers of Qwen3-30B-A3B (Figure~\ref{fig:sim}), highlighting their dominant role in shaping representations, while the later layers exhibit comparatively limited drift. We further examine the importance of accurate expert selection in these layers in Section~\ref{sec:imp_pf}.


\begin{figure}[h]
    \centering
    \includegraphics[width=1.0\linewidth]{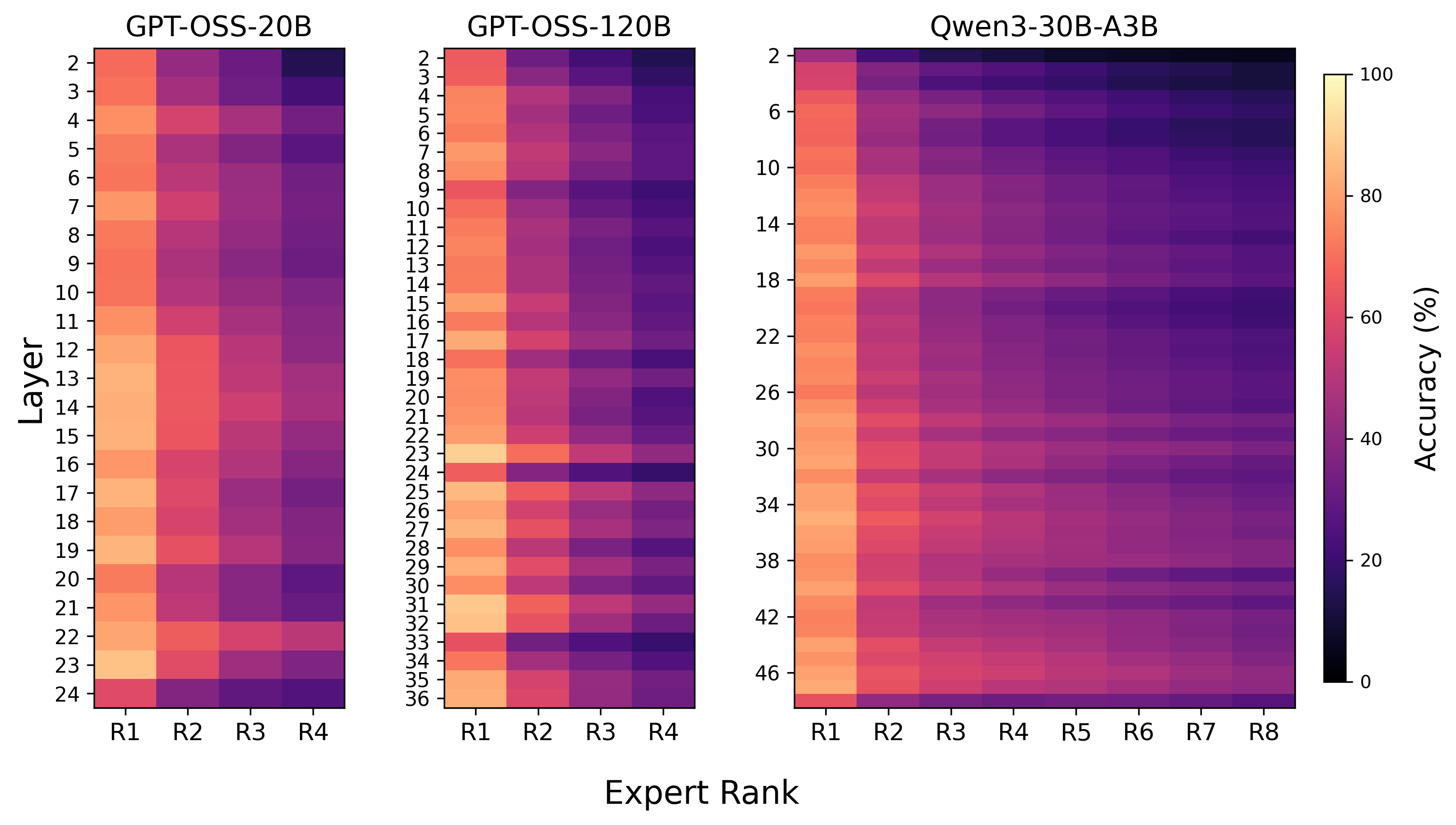}    
    \caption{Per-layer expert rank alignment between speculative and ground-truth routing. For each layer, we report the proportion for which the expert at a given rank (determined by its routing weight) matches between prefetched and ground-truth routing.}
    \label{fig:exp_rank_align}
\end{figure}

Expert rank alignment as shown in Figure~\ref{fig:exp_rank_align} provides an execution-level view of speculative routing and its implications for performance stability. We observe that the highest-ranked experts, i.e., those receiving the largest routing weights and contributing most to the MoE activation, are predicted with high hit rates across most layers. For Qwen3-30B-A3B, early layers and the final layer have considerably lower hit rates across all ranks. Low prediction hit rates are present primarily in lower-ranked experts with small probability and weighting mass, and whose contributions have limited impact on the resulting activation. This preservation of dominant experts helps to explain why downstream task accuracy remains stable despite imperfect routing predictions and suggests that the quasi-hidden state encodes sufficient information to recover the most influential experts for future layers. 

In the next section, we detail how we implement this prefetching scheme into an optimized inference engine. Based on this analysis, we report improvements using the speculated experts and do not re-fetch the true experts.

\section{Accelerating Inference}
Auto-regressive LLM inference has two phases: \emph{prefill}, followed by \emph{decode}. In prefill, the prompt tokens are processed in parallel
to generate the first output token. In decode, the model processes the previously generated token to predict the next token. As models scale, the memory footprint of parameters, activations, and the KV
cache can exceed the HBM capacity of a single GPU. Having multiple GPUs enables model sharding~\citep{megatronlm,huang2019gpipe}; however, our work focuses on resource-constrained deployment settings. In such scenarios, parameters or activations are often offloaded to CPU memory and transferred to
the GPU on demand, placing CPU$\rightarrow$GPU transfers on the critical path
and making them a key performance bottleneck. 

In this section, we analyze the effectiveness of prefetching across different inference phases. We demonstrate that it reduces the time per output token compared to on-demand loading of active experts, owing to the compute-memory overlap. 

\subsection{Decode Phase Prefetching}
\begin{algorithm}[t]
\caption{MoE Block with Expert Prefetching (Decode)}
\label{alg:moe-prefetch}
\begin{algorithmic}[1]
\REQUIRE Model $\mathcal{M}$; input activations $X \in \mathbb{R}^{B \times H}$ at layer $l$;
CPU-resident expert weights $\{W_{l,e}\}_{e=1}^{E}$; GPU expert cache/buffers $\mathcal{C}_l$;
(optional) prefetched indices $\mathcal{E}_l$ and gating weights $\mathcal{G}_l$
\ENSURE Output activations $O$; prefetched indices $\mathcal{E}_{l+1}$ and mixing weights $\mathcal{G}_{l+1}$
\IF{$l = 0$}
    \STATE \COMMENT{Cold start: No prefetched experts}
    \STATE $ids, g\gets \textsc{TopK}(\textsc{Gate}_{l}(X), k)$
    \STATE \textsc{CopySync}$(\mathcal{C}_l, \{W_{l,e}\}_{e \in ids})$ \COMMENT{CPU$\rightarrow$GPU blocking}
\ELSE
    \STATE $ids, g \gets \mathcal{E}_{l}, \mathcal{G}_{l}$
\ENDIF

\STATE $\mathcal{E}_{l+1}, \mathcal{G}_{l+1} \gets \textsc{GetNextExperts}(\textsc{Gate}_{l}(X), l)$
\STATE \textsc{WaitAndPrefetch}$(\mathcal{C}_{l},\mathcal{C}_{l+1}, \{W_{l+1, e}\}_{e \in \mathcal{E}_{l+1}})$ \COMMENT{CPU$\rightarrow$GPU non-blocking}
\STATE $O \gets \textsc{MoE}_{l}(X, ids, g, \mathcal{C}_{l})$ \COMMENT{runs on compute stream}

\end{algorithmic}
\end{algorithm}
Consider inference on a model $\mathcal{M}$ with $L$ layers, $E$ experts per MoE layer and $k$ experts per token, given a batch of $B$ prompts. We first focus on the decode phase, where one token is processed per prompt at each generation step. For small batch sizes, the number of active experts at layer $l$ satisfies $|\mathcal{E}_l| < E$. With $B{=}1$, as in single-user local inference, exactly $k$ experts are active
per layer - the minimum that must be loaded or
prefetched. As $B \to \frac{E}{k}$, the active expert set
$|\mathcal{E}_l| \to E$, assuming no overlap between active experts across tokens. Consequently, prefetching is only non-trivial with small batch sizes ($B \ll E$), requiring careful expert selection. We therefore focus on this regime and assume $B{=}1$, where expert transfers are minimized.

First, we model a baseline scheme that loads
experts on demand, after the routing step. In this
case, expert weight transfers lie on the critical path of execution, and the decode time or time-per-output-token (TPOT) is given by:
\begin{align*}
T_{\text{decode}}^{\text{ond}}
= \sum_{l=1}^{L} \Big(
t_{\text{attn},l}
+ t_{\text{gate+top-}k,l}
+ t_{\text{copy}}\!\left(\mathcal{E}_{l}\right)
+ t_{\text{expert},l}
\Big),
\end{align*}
where $t_{\text{attn},l}$ is the time for attention,
$t_{\text{gate+top-}k,l}$ is the time for routing and selecting the top-$k$ experts,
$t_{\text{copy}}\!\left(\mathcal{E}_{l}\right)$ is the time
to transfer the active experts to GPU memory, and $t_{\text{expert},l}$ is the time to compute the expert FFN. 

With prefetching, the transfer of next-layer expert weights is overlapped with computation from other operations, reducing the time on the critical path. This is given by:
\begin{align*}
T_{\text{decode}}^{\text{pf}}
= \sum_{l=1}^{L} \Big(
\max\!\big(
t_{\text{attn},l}
+ t_{\text{gate+top-}k,l}
+ t_{\text{expert},l},\\
\; t_{\text{copy}}\!\left(\mathcal{E}_{l+1}\right)
\big)
\Big)
\end{align*}
Let
$t_{\text{compute},l} =
t_{\text{attn},l} + t_{\text{gate+top-}k,l} + t_{\text{expert},l}$.
We can assume that
$t_{\text{copy}}\!\left(\mathcal{E}_{l}\right) =
t_{\text{copy}}\!\left(\mathcal{E}_{l+1}\right)$, as the
number of experts copied is the same for both layers. Therefore, the TPOT
improvement is given by:
\begin{align}
  \Delta T  &= \sum_{l=1}^{L}
  \min\!\left(
  t_{\text{copy}}\!\left(\mathcal{E}_{l}\right),
  \;
  t_{\text{compute},l}
  \right)
  \label{eq:improvement}
  \end{align}
Therefore, the expected TPOT improvement is bounded by the minimum of the total
compute time and the CPU$\rightarrow$GPU copy time. The maximum achievable
speedup is $2\times$, when compute and copy times are equal. In
practice, GPUs are optimized for high compute throughput,
and CPU$\rightarrow$GPU transfers occur over relatively slower
interconnects such as PCIe. As a result, copy time typically dominates
($t_{\text{copy}} \gg t_{\text{compute}}$), limiting the achievable speedup.
\begin{figure}[t]
\centering
\begin{lstlisting}[style=py]
def wait_and_prefetch(self, layer_idx, ids, next_ids):
    compute_stream = torch.cuda.current_stream()
 
    # Synchronize copy stream at layer 0
    if layer_idx == 0:
        self.copy_stream.wait_stream(
        compute_stream)

    # Wait for the current layer experts to finish copying
    self.wait_for_layer(layer_idx)

    # Load current experts if not resident
    if layer_idx not in self.layers_on_gpu:
        row_ids = ids.cpu()
        self.async_copy(layer_idx, row_ids)
        compute_stream.wait_stream(
        self.copy_stream)

    # Prefetch experts for the next layer
    if layer_idx + 1 < self.num_layers:
        row_ids = next_ids.cpu()
        self.async_copy(layer_idx+1, row_ids)

    # Yield control to overlap with compute
    yield

    # Offload previous layer
    self.layers_on_gpu.discard(layer_idx-1)
\end{lstlisting}
\caption{Layer-wise wait and prefetch code snippet.}
\label{fig:prefetch_forward}
\end{figure}

\noindent \textbf{Implementation Details.} 
We implement CPU offloading with both prefetching and on-demand expert loading in YALIS~\cite{yalis}, a light-weight research inference engine that supports key optimizations from SOTA frameworks such as
vLLM~\cite{kwon2023efficient}, including \texttt{torch.compile}, CUDA Graphs, and optimized attention backends.

Algorithm~\ref{alg:moe-prefetch} outlines the MoE block execution with expert prefetching. For the first layer, no prefetched experts are available, requiring synchronous CPU$\rightarrow$GPU copy of the active experts after routing and top-$k$ selection (Lines~1-3). For subsequent layers, we directly reuse the prefetched
expert indices and mixing weights (Lines~4-6). We then compute the next layer expert indices and mixing weights using our router-based scheme (Line 7). Thereafter, we wait for the current layer expert weights to be copied to the GPU, and initiate a transfer for the next layer experts (Line 8).
Due to hardware constraints, concurrent CPU$\rightarrow$GPU transfers are serialized to use the same copy engine and cannot be overlapped with each other. However, the asynchronous prefetch of the next layer experts overlaps with the execution of the current layer MoE computation(Line~9). This wait-and-prefetch logic is implemented as a Python context manager, and Figure~\ref{fig:prefetch_forward} provides a simplified
implementation snippet. We use double buffering to alternate GPU expert buffers across layers, enabling compute–copy overlap without additional synchronization. 

The offloaded MoE weights are stored in \emph{pinned} CPU memory, which cannot be paged out by the operating system and enables faster
CPU$\rightarrow$GPU transfers than pageable memory. We offload only the expert weights. Non-expert parameters (attention and router)
are small enough to remain resident on the GPU, avoiding unnecessary transfers.

\subsubsection{Experimental Setup}
In this section, we describe the experimental setup used to evaluate the
performance of decode-phase prefetching.

\begin{table}[h]
  \centering
  \caption{MoE model configuration and parameter memory (bf16). $L =$
  number of layers, $E = $ the number of experts per MoE layer, $H = $ model
  hidden size, $H_{\text{MoE}} = $ expert hidden size. $M_{\text{experts}}$
  and $M_{\text{other}}$ denote the memory footprint of the experts and other
  parameters, respectively.}
  \label{tab:model_config}
  \setlength{\tabcolsep}{4pt} 
  \resizebox{\columnwidth}{!}{%
    \definecolor{lightgrey}{gray}{0.92}
  \rowcolors{2}{lightgrey}{white}
  \begin{tabular}{l|c|c|c|c|c|c}
  \toprule
  Model & $L$ & $E$ & $H$ & $H_{\text{MoE}}$ & $M_{\text{experts}}$ & $M_{\text{other}}$ \\
  \midrule
  Qwen3-30B-A3B & 48 & 128 & 2048 & 768 & $\sim$54\,GB & $\sim$3\,GB \\
  GLM-4.7-Flash & 47 & 64 & 2048 & 1536 & $\sim$53\,GB & $\sim$3\,GB \\
  GPT-OSS-120B & 24 & 32 & 2880 & 2880 & $\sim$213\,GB & $\sim$4\,GB \\
  Qwen3-235B-A22B & 94 & 128 & 4096 & 1536 & $\sim$230\,GB & $\sim$14\,GB \\
  \bottomrule
  \end{tabular}
  }
\end{table}

\noindent \textbf{Models.}
Table~\ref{tab:model_config}
lists the models used in our evaluation and their associated parameter memory costs. For GLM-4.7-Flash and GPT-OSS-120B, we retain the MoE layer configurations, but use random expert weights and a simplified RoPE-based attention mechanism due to limitations in YALIS. For
GLM-4.7-Flash, we use only the routed experts and discard the shared expert
due to incompatibility with the inference engine. The diverse set of models chosen represents the general applicability of our scheme to various MoE configurations. All evaluations use \texttt{bf16} precision. 

\noindent \textbf{Method and Baselines} We run both prefetching and on-demand expert loading with YALIS. For prefetching, we use the router-based scheme described in Section~\ref{sec:effec_pfing}.
We offload only the expert weights to CPU memory, keeping the router and attention parameters on the GPU due to their small memory footprint (Table~\ref{tab:model_config}).

\begin{table}[h]

\centering
\caption{Hardware configurations used for evaluation.}
\label{tab:hardware}
\setlength{\tabcolsep}{4pt}
\resizebox{\columnwidth}{!}{%
  \definecolor{lightgrey}{gray}{0.92}
  \rowcolors{2}{lightgrey}{white}
\begin{tabular}{lcccp{3.5cm}}
\toprule
GPU & HBM & CPU DRAM & CPU$\leftrightarrow$GPU Link& Models Evaluated \\
\midrule
A6000 & 48\,GB & 128\,GB & PCIe 4.0 &
Qwen3-30B-A3B, GLM-4.7-Flash \\
A100 & 80\,GB & 256\,GB & PCIe 4.0 &
GPT-OSS-120B \\
GH200 & 96\,GB & 480\,GB & NVLink C2C &
Qwen3-235B-A22B \\
\bottomrule
\end{tabular}
  }
\end{table}
\noindent \textbf{Hardware and Environment.}
Table~\ref{tab:hardware} describes the hardware configurations used in our
evaluation. In all settings, model parameters exceed available GPU memory,
necessitating CPU offloading. We use PyTorch~2.9 and the
FlashAttention backend~\citep{dao2022flashattention}. For performance analysis, we collect execution traces using Nsight Systems~\citep{nsys} and analyze them with Pipit~\citep{bhatele:2023pipit}. Although our experiments use NVIDIA GPUs, both YALIS and our prefetching scheme are portable to AMD, which we plan to evaluate in future work.

\noindent \textbf{Workload and Metrics.} 
With a batch size of one, we vary the prompt length over $\{1024, 4096, 16384, 65536\}$ and generate 32 tokens. We report time-per-output-token (TPOT), defined as the average time to generate each output token after the first. Decode computation scales with context length. Varying the prompt length thus directly controls the decode compute. The small number of output tokens provides stable averaging without materially changing the per-token compute. We average results over three generations and three independent trials and report error bars.

\subsubsection{On-Demand Loading Performance Breakdown}

\begin{figure}[t]
    \centering
    \includegraphics[width=\columnwidth]{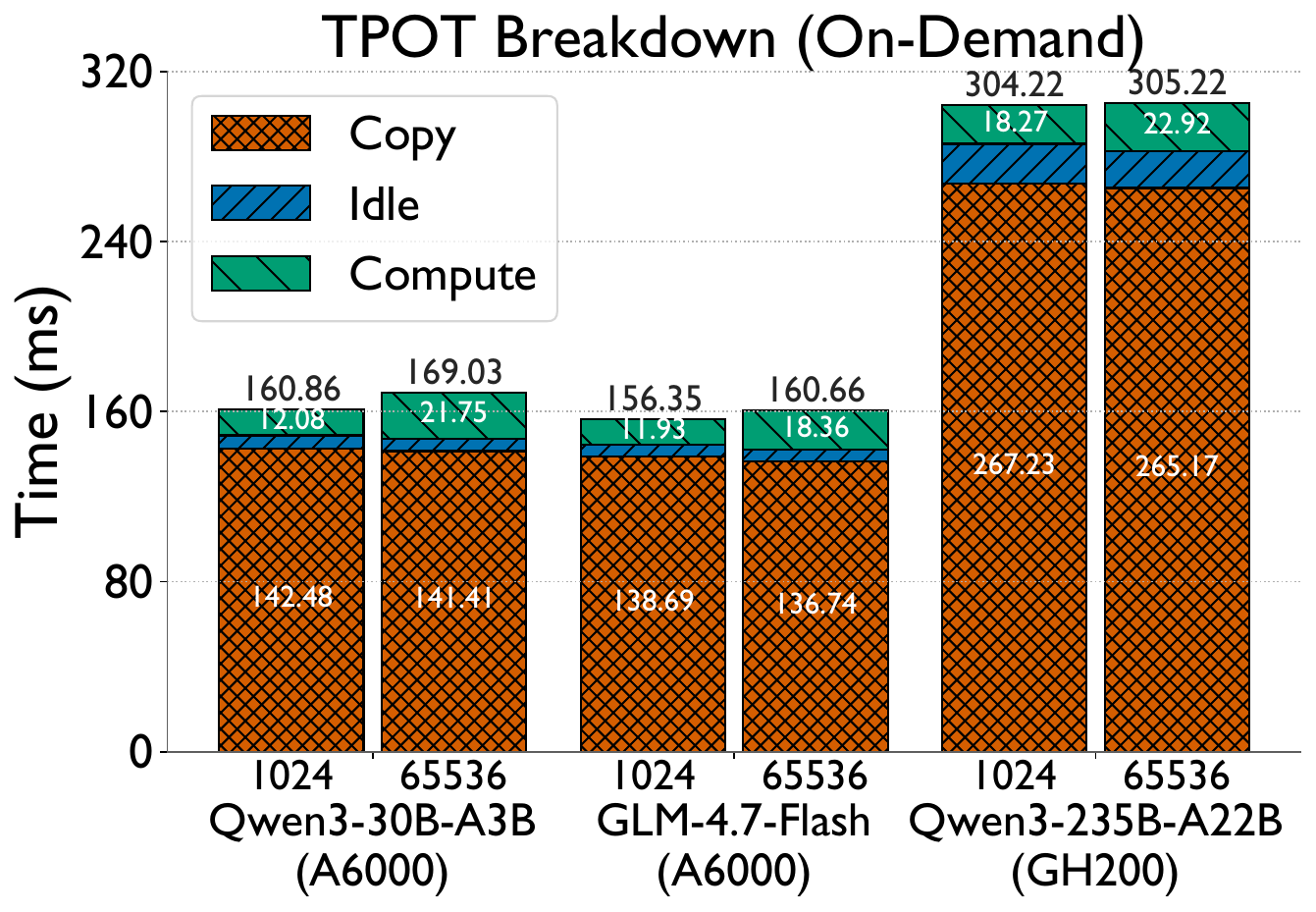}
    \caption{Breakdown of time per output token for on-demand expert loading across various models and prompt sequence lengths.}
    \label{fig:breakdown}
\end{figure}

We first examine the breakdown of TPOT under on-demand expert loading into \emph{compute}, \emph{copy}, and \emph{idle} times (Figure~\ref{fig:breakdown}). Idle time corresponds to periods when the GPU is not actively executing a kernel or a memory copy, and includes launch and synchronization overheads. 

Focusing on Qwen3-30B-A3B, we observe that copy time constitutes the primary bottleneck, accounting for $\sim$84–88\% of TPOT. This trend is consistent across all models studied. Compute time accounts for  $\sim$8–13\% of the total TPOT, which forms an upper bound on the achievable improvement (Equation~\ref{eq:improvement}). In the following section, we demonstrate that our implementation achieves this bound.  

Comparing context lengths 1024 and 65536, we observe that copy times remain constant, but compute time increases with context length for all models. This trend indicates that overlapping compute with copy through prefetching yields greater benefits at longer sequence lengths. Due to the lack of KV-cache offloading support in YALIS, we were limited by GPU HBM capacity and unable to run even larger sequence lengths.

\subsubsection{Comparing On-Demand Loading Vs. Prefetching}
\begin{figure*}[t]
    \centering
    \includegraphics[height=1.235in]{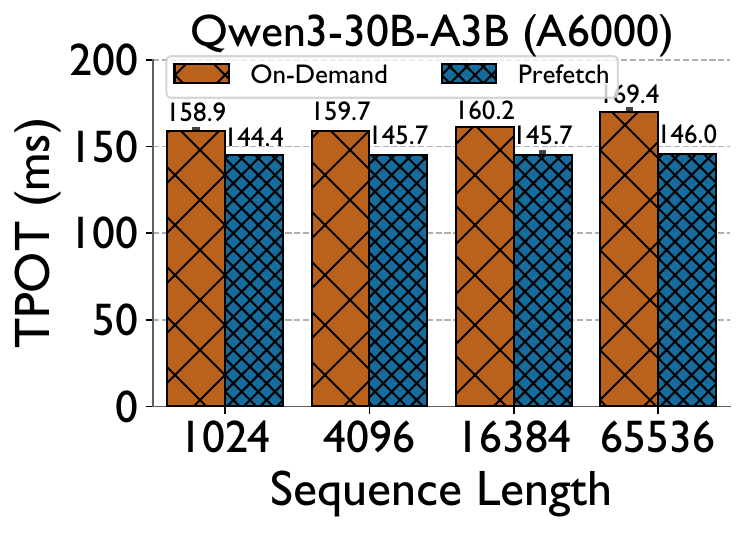}
    \includegraphics[height=1.235in]{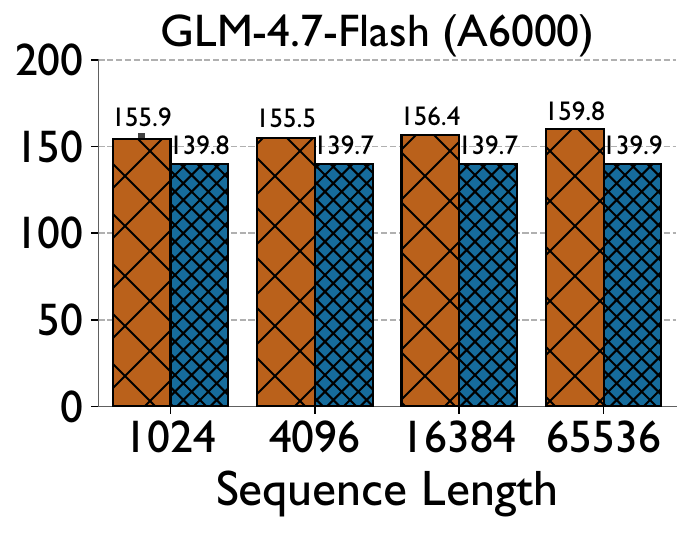}
    \includegraphics[height=1.235in]{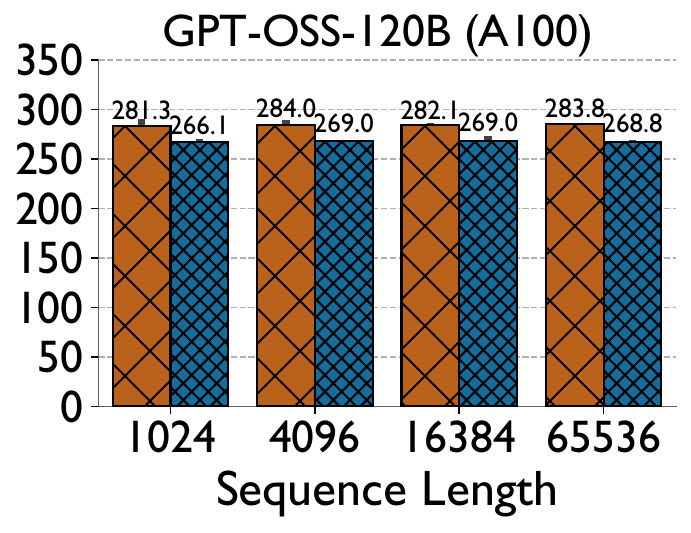}
    \includegraphics[height=1.235in]{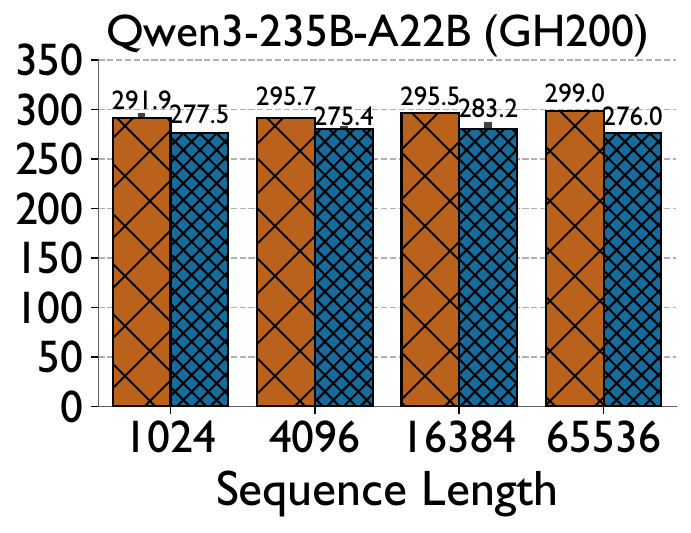}
    \caption{Time per output token (TPOT) for different models, comparing on-demand expert loading and expert prefetching across varying sequence lengths.}
    \label{fig:tbt_comparison}
\end{figure*}

Figure~\ref{fig:tbt_comparison} compares the TPOT of on-demand expert loading with our prefetching scheme across different models and prompt sequence lengths. For Qwen3-30B-A3B, prefetching yields a 9–14\% reduction in TPOT, with bigger gains at longer sequence lengths. This behavior is consistent with the breakdowns in Figure~\ref{fig:breakdown} and indicates that our implementation closely approaches the maximum achievable speedup for the observed compute–copy time split. Similar trends are observed across all evaluated models and architectures. 

On more powerful GPUs (A100 and GH200; right two plots), the maximum TPOT improvement is limited to 5–8\%, compared to 12–14\% on the A6000. This difference is primarily due to the higher compute throughput of these GPUs, indicating that prefetching provides larger gains on weaker GPUs. Overall, our prefetching approach yields performance improvements
across diverse MoE models and hardware configurations. These performance improvements arise from overlapping computation and data
transfers on separate CUDA streams, as illustrated by the Nsight Systems traces in Figure~\ref{fig:trace} for the Qwen3-30B-A3B model (A6000, context length 65536).

\subsection{Prefill Phase Prefetching}
We now consider the prefill phase, in which all $p_i$ prompt tokens $(0 \le i < B)$ are processed in parallel. For typical inference workloads, prompt lengths are sufficiently large that all experts are effectively active at every layer, even when $B = 1$. Consequently, any expert prediction or prefetching strategy trivially reduces to loading all experts during prefill. While prefetching still outperforms on-demand loading by overlapping computation with data transfers, it does not involve any non-trivial expert selection decisions in this regime, as shown in prior work~\citep{zhang25_duoserve_moe_expert_pf}. Therefore, although we implement this strategy in our prefetching system, we do not analyze the prefill regime further. 






\section{Improving Prefetching Accuracy}
\label{sec:imp_pf}

The quasi-hidden state provides a strong predictive signal, closely aligning with the input anticipated by the learned router function. For architectures such as GPT-OSS, this alignment is sufficiently strong, enabling accurate expert speculation throughout the network and preserving downstream task performance. 

With Qwen3-30B-A3B, the early layers have higher representational drift, indicating a regime of increased uncertainty in the hidden states. In this regime, the mismatch between speculated experts and the router's ground truth selections is substantially higher (as seen in Figure~\ref{fig:hit_rates}). Errors in expert prediction at these earlier layers may perturb downstream token representations significantly, resulting in a noticeable degradation in task performance. This suggests that improving expert prediction hit rate across layers (or specifically layers where there is high representational uncertainty) can mitigate the downstream performance loss under speculative execution. 

\subsection{Neural Expert Prefetcher}

Motivated by this, we train a lightweight neural estimator that directly predicts routing decisions from layer-$l$ representations. The estimator learns a mapping from the quasi-hidden state $q_l$ to the router logits at layer $l+1$. The estimator is designed to be shallow to avoid additional serial bottleneck and incur minimal runtime overhead, making it suitable as a drop-in replacement for router execution during inference.  

Architecturally, it is a feed-forward network with a learned positional embedding to account for layer-specific routing behavior (see Appendix~\ref{arch} for more information). Training is performed via distillation, minimizing the Kullback-Leibler divergence between the estimator's predicted logits and those produced by the ground-truth router. We train estimators with the following parameter counts: 4M for GPT-OSS-20B, 45M parameters for GPT-OSS-120B, and 17M parameters for Qwen3-30B-A3B.

\subsection{Training and Hit Rate Improvements}

\begin{figure}[H]
    \centering
    \includegraphics[width=0.9\linewidth]{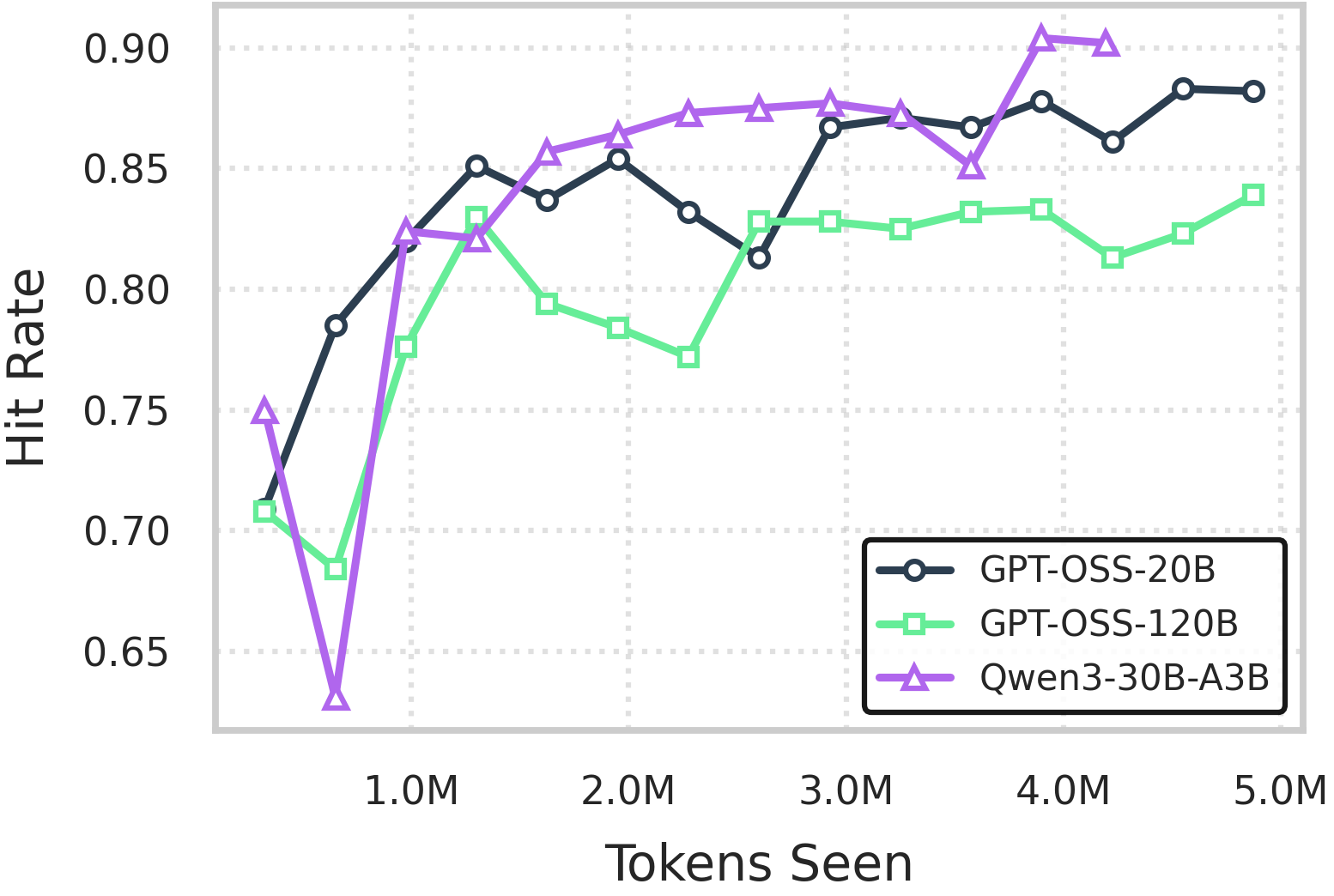}    
    \caption{Validation expert hit rate versus the number of tokens used to train the expert estimator.}
    \label{fig:tokens_hr}
\end{figure}

We find that the estimators achieve high prediction accuracy with relatively few training tokens (Figure~\ref{fig:tokens_hr}). After 4M tokens, the estimator for Qwen3-30B-A3B reaches an average hit rate of approximately 90\%. After 5M tokens, the estimators for GPT-OSS-120B and GPT-OSS-20B achieve average hit rates of 83\% and 88\%, respectively, with additional training yielding diminishing returns.

\begin{figure}[h]
    \centering
    \includegraphics[width=0.9\linewidth]{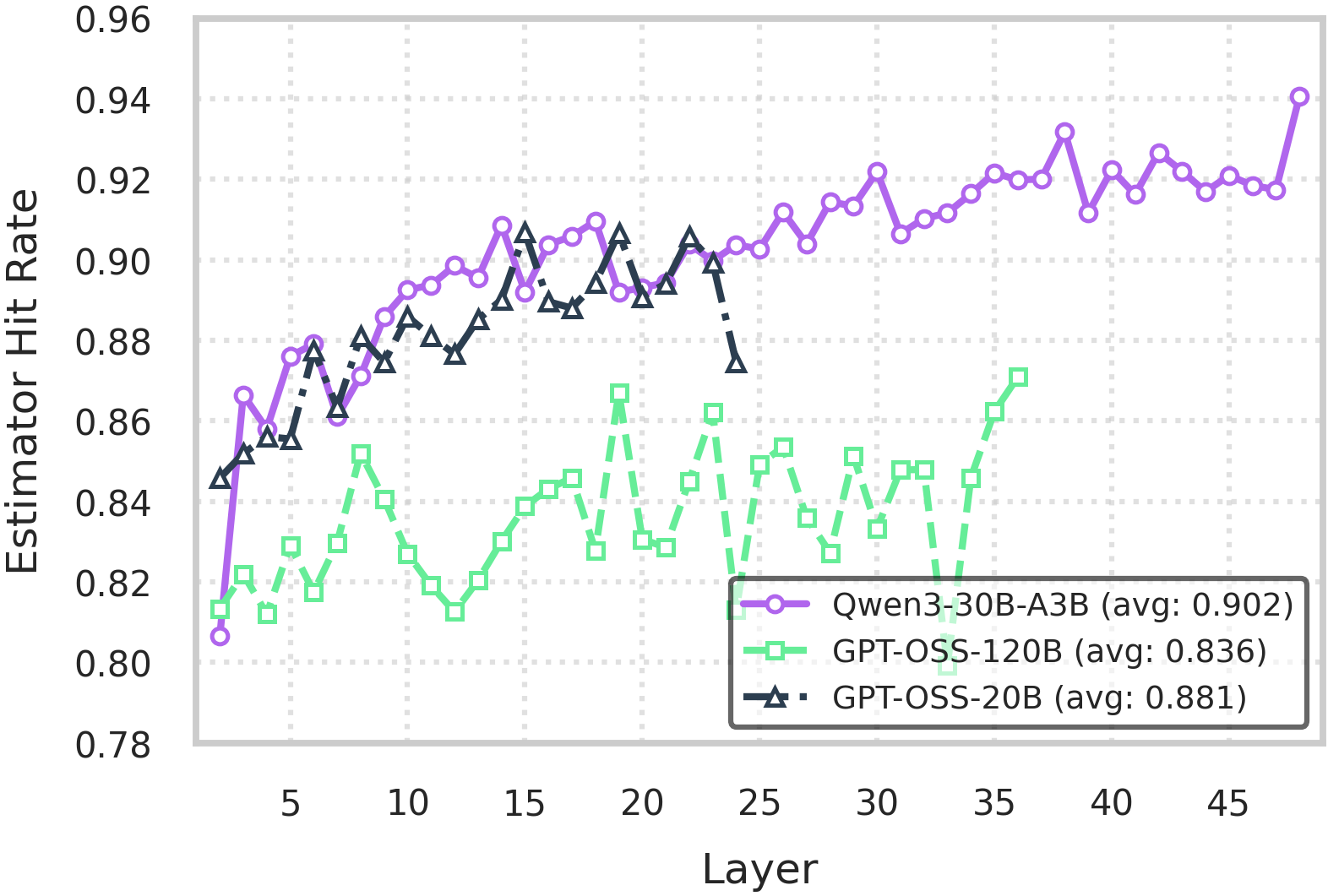}    
    \caption{Per-layer expert hit rate of model-specific trained estimators measured against ground-truth router-selected experts.}
    \label{fig:est_hr}
\end{figure}

Figure~\ref{fig:est_hr} reports the per-layer hit rates for the trained estimators. For Qwen3-30B-A3B, the estimator significantly improves expert prediction accuracy in the early layers compared to the router-based speculative execution, with a maximum improvement of approximately 25\% over using the fixed router.

\subsection{Estimator and Hybrid Prefetching}
We evaluate two estimator-based inference strategies as noted in Table~\ref{tab:model_performance}. In the first, denoted \textbf{Est-PF}, the estimator replaces router-based expert predictions at all layers. In the second, denoted \textbf{Hybrid-PF}, the estimator is applied to layers where router-based hit rates are low, while quasi-hidden state based predictions are used everywhere else. 

The results in Table~\ref{tab:model_performance} show that for the GPT-OSS models, estimator-based approaches offer limited additional benefit in general. This is expected since router-based speculative execution already preserves downstream task accuracy for these architectures. However, for Qwen3-30B-A3B, the \textbf{Est-PF} scheme substantially improves performance across all evaluated benchmarks relative to router-based prefetching.

Importantly, the \textbf{Hybrid-PF} strategy recovers most of the baseline performance for Qwen3-30B-A3B while applying the estimator only where prefetch hit rates were low. Replacing expert prediction in early layers alone is sufficient to significantly improve downstream task accuracy, most notably on GSM8k, where the hybrid approach recovers approximately 37\% of the accuracy gap to \textbf{Router-PF}, bringing performance substantially closer to the baseline. This supports the hypothesis that early layer expert mismatch significantly influenced the degradation in accuracy. 

\subsection{Additional Implications}
Since the estimator directly predicts the router logits a layer in advance, router execution is not strictly required at inference time once the $s_l$ representation is available. This provides a router-free inference path where subsequent routing decisions are produced by the estimator, eliminating repeated router calls and their serial overhead. While our experiments focus on \textbf{Est-PF} and \textbf{Hybrid-PF} for accuracy preservation, this direction may improve parallelism.

\section{Conclusion}
\label{sec:conclusion}

MoEs enable scaling to hundreds of billions of parameters via sparse activations \cite{switch_transformer}. Under memory-constrained settings that rely on offloading, inference becomes I/O-bound due to CPU-GPU expert transfers, reducing TPOT. We introduce an inference-time expert prefetching scheme that speculates future expert selection from internal model representations, enabling expert transfers to overlap with computation and improving TPOT. Our parameter-free approach uses the quasi-hidden state to predict next-layer experts over several modern MoE architectures, while a lightweight estimator selectively improves hit rate over high drift layers. Integrated into the YALIS inference engine, our approach achieves up to 5-14\% TPOT improvement over on-demand expert loading while largely preserving downstream task accuracy. By reducing the cost incurred by expert offloading, this work makes large open-source MoE models more practical for local deployment on consumer hardware. 

\section{Future Work}
\label{sec:future_work}
Our analysis of prefetching focused primarily on CPU-GPU expert offloading settings on consumer and single-GPU datacenter hardware. A natural extension is to study more resource-constrained deployments (e.g., smartphones, robotic platforms, and embedded systems), where prefetching may be applied to offloading involving disk-CPU memory transfers. Another direction is further examination of router-free inference: replacing per-layer routing with a single trained estimator, which we have shown enables prefetching with minimal quality degradation. Validation in broader settings should explore tasks beyond reasoning and focus on integration into optimized inference paths. Finally, motivated by the representational similarities observed across many layers, multi-layer prefetching may further improve compute-memory overlap in conjunction with on-device expert caching.

\section*{Impact Statement}

This paper presents work whose goal is to advance the field of Machine
Learning. There are many potential societal consequences of our work, none
which we feel must be specifically highlighted here.


\bibliographystyle{icml2026}
\bibliography{main}

@misc{chen25_spmoe,
      title={SP-MoE: Speculative Decoding and Prefetching for Accelerating MoE-based Model Inference}, 
      author={Liangkun Chen and Zijian Wen and Tian Wu and Xiaoxi Zhang and Chuan Wu},
      year={2025},
      eprint={2510.10302},
      archivePrefix={arXiv},
      primaryClass={cs.DC},
      url={https://arxiv.org/abs/2510.10302}, 
    }

@misc{hwang24_pregated_moe,
      title={Pre-gated MoE: An Algorithm-System Co-Design for Fast and Scalable Mixture-of-Expert Inference}, 
      author={Ranggi Hwang and Jianyu Wei and Shijie Cao and Changho Hwang and Xiaohu Tang and Ting Cao and Mao Yang},
      year={2024},
      eprint={2308.12066},
      archivePrefix={arXiv},
      primaryClass={cs.LG},
      url={https://arxiv.org/abs/2308.12066}, 
}

@misc{zhang25_duoserve_moe_expert_pf,
      title={DuoServe-MoE: Dual-Phase Expert Prefetch and Cache Scheduling for Efficient MoE LLM Inference}, 
      author={Yuning Zhang and Grant Pinkert and Nan Yang and Yanli Li and Dong Yuan},
      year={2025},
      eprint={2509.07379},
      archivePrefix={arXiv},
      primaryClass={cs.DC},
      url={https://arxiv.org/abs/2509.07379}, 
}

@misc{yany25_qwen3_tech_report,
      title={Qwen3 Technical Report}, 
      author={Qwen3 Team},
      year={2025},
      eprint={2505.09388},
      archivePrefix={arXiv},
      primaryClass={cs.CL},
      url={https://arxiv.org/abs/2505.09388}, 
}

@misc{openai25_oss_model_card,
      title={gpt-oss-120b and gpt-oss-20b Model Card}, 
      author={{OpenAI}},
      year={2025},
      eprint={2508.10925},
      archivePrefix={arXiv},
      primaryClass={cs.CL},
      url={https://arxiv.org/abs/2508.10925}, 
}

@misc{glm45_25_tech_report,
      title={{GLM-4.5}: Agentic, Reasoning, and Coding (ARC) Foundation Models}, 
      author={ {GLM-4.5 Team}},
      year={2025},
      eprint={2508.06471},
      archivePrefix={arXiv},
      primaryClass={cs.CL},
      url={https://arxiv.org/abs/2508.06471}, 
}

@misc{yu25_taming_latency_mem_tradeoff_exp_offloading,
      title={Taming Latency-Memory Trade-Off in MoE-Based LLM Serving via Fine-Grained Expert Offloading}, 
      author={Hanfei Yu and Xingqi Cui and Hong Zhang and Hao Wang and Hao Wang},
      year={2025},
      eprint={2502.05370},
      archivePrefix={arXiv},
      primaryClass={cs.LG},
      url={https://arxiv.org/abs/2502.05370}, 
}

@misc{xue2025_moe_infinity,
      title={MoE-Infinity: Efficient MoE Inference on Personal Machines with Sparsity-Aware Expert Cache}, 
      author={Leyang Xue and Yao Fu and Zhan Lu and Luo Mai and Mahesh Marina},
      year={2025},
      eprint={2401.14361},
      archivePrefix={arXiv},
      primaryClass={cs.LG},
      url={https://arxiv.org/abs/2401.14361}, 
}

@misc{yu25_pre_scope,
      title={PreScope: Unleashing the Power of Prefetching for Resource-Constrained MoE Inference}, 
      author={Enda Yu and Zhaoning Zhang and Dezun Dong and Yongwei Wu and Xiangke Liao},
      year={2025},
      eprint={2509.23638},
      archivePrefix={arXiv},
      primaryClass={cs.LG},
      url={https://arxiv.org/abs/2509.23638}, 
}

@inproceedings{large_nns,title	= {Outrageously Large Neural Networks: The Sparsely-Gated Mixture-of-Experts Layer},author	= {Noam Shazeer and Azalia Mirhoseini and Krzysztof Maziarz and Andy Davis and Quoc Le and Geoffrey Hinton and Jeff Dean},year	= {2017},URL	= {https://openreview.net/pdf?id=B1ckMDqlg}}

@misc{panda2025densebackpropagationimprovestraining,
      title={Dense Backpropagation Improves Training for Sparse Mixture-of-Experts}, 
      author={Ashwinee Panda and Vatsal Baherwani and Zain Sarwar and Benjamin Therien and Sambit Sahu and Tom Goldstein and Supriyo Chakraborty},
      year={2025},
      eprint={2504.12463},
      archivePrefix={arXiv},
      primaryClass={cs.LG},
      url={https://arxiv.org/abs/2504.12463}, 
}

@misc{eliseev_fast_inference,
      title={Fast Inference of Mixture-of-Experts Language Models with Offloading}, 
      author={Artyom Eliseev and Denis Mazur},
      year={2023},
      eprint={2312.17238},
      archivePrefix={arXiv},
      primaryClass={cs.LG},
      url={https://arxiv.org/abs/2312.17238}, 
}

@techreport{megatronlm,
  title         = {Megatron-LM: Training Multi-Billion Parameter Language Models Using Model Parallelism},
  author        = {Mohammad Shoeybi and Mostofa Patwary and Raul Puri and Patrick LeGresley and Jared Casper and Bryan Catanzaro},
  year          = {2020},
  eprint        = {1909.08053},
  archiveprefix = {arXiv},
  primaryclass  = {cs.CL}
}

@article{huang2019gpipe,
  title={Gpipe: Efficient training of giant neural networks using pipeline parallelism},
  author={Huang, Yanping and Cheng, Youlong and Bapna, Ankur and Firat, Orhan and Chen, Dehao and Chen, Mia and Lee, HyoukJoong and Ngiam, Jiquan and Le, Quoc V and Wu, Yonghui and others},
  journal={Advances in neural information processing systems},
  volume={32},
  year={2019}
}

@article{yalis,
  title={LLM Inference Beyond a Single Node: From Bottlenecks to Mitigations with Fast All-Reduce Communication},
  author={Singhania, Prajwal and Singh, Siddharth and Hough, Lannie Dalton and Srivastava, Akarsh and Menon, Harshitha and Jekel, Charles Fredrick and Bhatele, Abhinav},
  journal={arXiv preprint arXiv:2511.09557},
  year={2025}
}

@article{dao2022flashattention,
  title={Flashattention: Fast and memory-efficient exact attention with io-awareness},
  author={Dao, Tri and Fu, Dan and Ermon, Stefano and Rudra, Atri and R{\'e}, Christopher},
  journal={Advances in neural information processing systems},
  volume={35},
  pages={16344--16359},
  year={2022}
}

@misc{bhatele:2023pipit,
      title={Pipit: Scripting the analysis of parallel execution traces},
      author={Abhinav Bhatele and Rakrish Dhakal and Alexander Movsesyan and Aditya K. Ranjan and Onur Cankur},
      year={2023},
      eprint={2306.11177},
      archivePrefix={arXiv},
      primaryClass={cs.DC}
}

@misc{nsys,
  author       = {NVIDIA},
  title        = {NVIDIA Nsight Systems},
  year         = {2018},
  publisher    = {},
  journal      = {},
  howpublished = {\url{https://developer.nvidia.com/nsight-systems}},
  commit       = {}
}

@misc{switch_transformer,
      title={Switch Transformers: Scaling to Trillion Parameter Models with Simple and Efficient Sparsity}, 
      author={William Fedus and Barret Zoph and Noam Shazeer},
      year={2022},
      eprint={2101.03961},
      archivePrefix={arXiv},
      primaryClass={cs.LG},
      url={https://arxiv.org/abs/2101.03961}, 
}

@inproceedings{kwon2023efficient,
  title={Efficient memory management for large language model serving with pagedattention},
  author={Kwon, Woosuk and Li, Zhuohan and Zhuang, Siyuan and Sheng, Ying and Zheng, Lianmin and Yu, Cody Hao and Gonzalez, Joseph and Zhang, Hao and Stoica, Ion},
  booktitle={Proceedings of the 29th symposium on operating systems principles},
  pages={611--626},
  year={2023}
}

@misc{zhang25_ladder_res,
      title={Ladder-residual: parallelism-aware architecture for accelerating large model inference with communication overlapping}, 
      author={Muru Zhang and Mayank Mishra and Zhongzhu Zhou and William Brandon and Jue Wang and Yoon Kim and Jonathan Ragan-Kelley and Shuaiwen Leon Song and Ben Athiwaratkun and Tri Dao},
      year={2025},
      eprint={2501.06589},
      archivePrefix={arXiv},
      primaryClass={cs.LG},
      url={https://arxiv.org/abs/2501.06589}, 
}

\newpage
\appendix

\section{Estimator Architecture}
\label{arch}
\begin{figure}[h]
    \centering
    \includegraphics[scale=0.6]{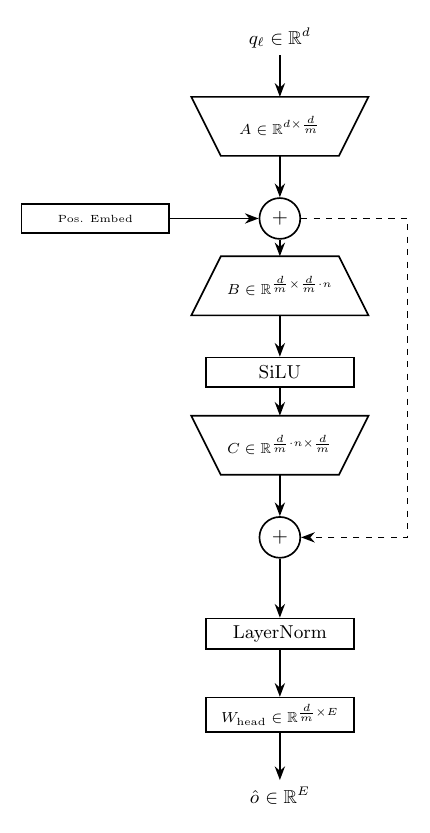}
    \caption{Outline of the neural estimator architecture. $m, n > 1$}
    \label{fig:model_diagram}
\end{figure}

Figure~\ref{fig:model_diagram} illustrates the straight forward architecture of the neural expert estimator. Given the quasi-hidden state $q_l \in \mathbb{R}^d$ at layer $l$, the estimator first applies a linear projection $A \in \mathbb{R}^{d \times \frac{d}{m}}$ to obtain a reduced latent representation. A learned positional embedding corresponding to layer $l$ is then added to this representation to encode layer-specific routing behavior. The resulting vector is passed through a feed-forward network consisting of two linear layers $B \in \mathbb{R}^{\frac{d}{m} \times \frac{d}{m}\cdot n}$ and $C \in \mathbb{R}^{\frac{d}{m}\cdot n \times \frac{d}{m}}$ with a SiLU non-linearity. A residual connection skips over the MLP block, and the output is normalized using a LayerNorm. Finally, a linear head $W_{\text{head}} \in \mathbb{R}^{\frac{d}{m} \times E}$ produces predicted router logits $\hat{o} \in \mathbb{R}^E$ for layer $l+1$, where $E$ denotes the number of experts.

\onecolumn
% ...

\end{document}